\documentclass[runningheads]{llncs}
\usepackage{graphicx}
\usepackage[babel]{csquotes}
\usepackage{enumitem}
\usepackage[hyphens]{url}
\usepackage{amsmath}
\usepackage{amsfonts}
\usepackage{tabularx}
\usepackage{xcolor}
\usepackage[linesnumbered]{algorithm2e}
\SetAlCapSty{}
\SetAlgoInsideSkip{bigskip}
\usepackage{hyperref}

\newcommand{\labs}{\left|}
\newcommand{\rabs}{\right|}

\begin{document}
\title{Employing an Adjusted Stability Measure for Multi-Criteria Model Fitting on Data Sets with Similar Features}
\titlerunning{Employing an Adjusted Stability Measure}
%
\author{Andrea Bommert \and
J\"org Rahnenf\"uhrer \and
Michel Lang}
\authorrunning{A. Bommert et al.}
%
\institute{Department of Statistics, TU Dortmund University, 44221 Dortmund, Germany
\email{\{bommert,rahnenfuehrer,lang\}@statistik.tu-dortmund.de}\footnote{The R source code for all analyses  presented in this paper is publicly available at \url{https://github.com/bommert/model-fitting-similar-features}}}
\maketitle              
\begin{abstract}
Fitting models with high predictive accuracy that include all relevant but no irrelevant or redundant features is a challenging task on data sets with similar (e.g.\ highly correlated) features.
We propose the approach of tuning the hyperparameters of a predictive model in a multi-criteria fashion with respect to predictive accuracy and feature selection stability.
We evaluate this approach based on both simulated and real data sets and we compare it to the standard approach of single-criteria tuning of the hyperparameters as well as to the state-of-the-art technique \enquote{stability selection}.
We conclude that our approach achieves the same or better predictive performance compared to the two established approaches.
Considering the stability during tuning does not decrease the predictive accuracy of the resulting models.
Our approach succeeds at selecting the relevant features while avoiding irrelevant or redundant features.
The single-criteria approach fails at avoiding irrelevant or redundant features and the stability selection approach fails at selecting enough relevant features for achieving acceptable predictive accuracy.
For our approach, for data sets with many similar features, the feature selection stability must be evaluated with an adjusted stability measure, that is, a measure that considers similarities between features.
For data sets with only few similar features, an unadjusted stability measure suffices and is faster to compute.
\end{abstract}

\section{Introduction}
Feature selection and model fitting are some of the most fundamental problems in data analysis, machine learning, and data mining.
Especially for high-dimensional data sets, it is often advantageous with respect to predictive performance, run time, and interpretability to disregard the irrelevant and redundant features.
This can be achieved by choosing a suitable subset of features that are relevant for target prediction.
The standard approach for fitting predictive models is tuning their hyperparameters only with respect to predictive accuracy.
However, multi-criteria tuning approaches have been applied successufully for obtaining models that do not only achieve high predictive accuracy but that are also sparse and have a stable feature selection~\cite{bommert2017multicriteria}.
The stability of a feature selection algorithm is defined as the robustness of the set of selected features towards different data sets from the same data generating distribution~\cite{kalousis2007stability}.
Stability quantifies how different training data sets affect the sets of selected features.

Many high-dimensional data sets contain similar features.
An example are gene expression data sets with genes of the same biological processes often being highly positively correlated.
For continuous features, the Pearson correlation is often used to quantify the similarity between features.
But other criteria, possibly also measuring non-linear associations, can be considered as well.
For categorical features, information theoretic quantities like mutual information can be employed.
For data sets with highly similar features, feature selection is very challenging, because it is more difficult to avoid the selection of relevant but redundant features.
For such data sets, also the evaluation of feature selection stability is more difficult.
Unadjusted stability measures see features with different identifiers as different features.
Consider a situation with one set containing a feature $X_A$ and another set not including $X_A$ but instead an almost identical feature $X_B$.
Even though $X_A$ and $X_B$ provide almost the same information, unadjusted measures consider the selection of $X_B$ instead of $X_A$ (or vice versa) as a lack of stability.
Adjusted stability measures on the other hand take into account the similarities between the features but require more time for calculation~\cite{bommert2020adjusted}.

Performing feature selection on data sets with similar features is not a prominent issue in the literature.
Some benchmark studies include scenarios with similar features, see for example \cite{dash1997feature,hall1999correlation}.
Also, feature selection methods for selecting relevant and avoiding redundant features have been defined, for example in \cite{brown2012conditional,yu2004efficient}.
These methods are greedy forward search algorithms that measure the redundancy of features by their similarity to the already selected features.
Such sequential search methods, however, are infeasible for high-dimensional data sets.
In accordance with \cite{brown2012conditional} and \cite{yu2004efficient}, we find it desirable to have only one feature per group of relevant and similar features included in the model.
This allows an easier interpretation because the model is more sparse.
In preliminary studies on data sets with highly similar features, we have observed that established feature selection methods, such as lasso regression or random forest, are not able to select only one feature out of a group of similar features.
Instead, they select several features out of groups of relevant and similar features, that is, they select many redundant features.
A method that is able to select only one feature per group of relevant and similar features is $L_0$-regularized regression.
For recent work on efficient computation of $L_0$-regularized regression see for example~\cite{hazimeh2018fast}.

In this paper, our goal is finding models with high predictive accuracy that include all relevant information for target prediction but no irrelevant or redundant features.
For achieving this for data sets with similar features, we propose considering both the predictive accuracy and the feature selection stability during hyperparameter tuning.
This idea has also been described in the first author's dissertation~\cite{bommert2020integration}.
Configurations with a stable feature selection select almost the same features for all data sets (here: cross-validation splits).
If the same features are selected for slightly varying data sets, these features are presumably relevant and not redundant.
If the feature selection stability is low, many features are only included in some of the models.
In this case, it is likely that these features are either redundant or do not carry much information for target prediction.
Therefore, considering the feature selection stability during hyperparameter tuning should lead to models that include neither irrelevant nor redundant features.
For data sets with similar features, it is expected that the selected features vary such that features with different identifiers but almost identical information are selected.
This is taken into account when employing an adjusted stability measure.
We compare our approach to competing approaches based on both simulated data and real data.
The aim of the analyses on simulated data is finding out whether the proposed approach allows fitting models that include all features that were used for target generation and no irrelevant or redundant features.
On real data, the features that generate the target variable are unknown, but the performance of the models with respect to predictive accuracy and number of selected features can be examined.

The remainder of the paper is organized as follows: 
In Section~\ref{sec.methods}, the proposed approach and competing approaches are explained in detail.
Comparative experiments based on simulated and on real data are conducted in Sections~\ref{sec.exp.sd} and~\ref{sec.exp.trd}, respectively.
Section~\ref{sec.conclusion} contains a summary and concluding remarks.

\section{Methods}\label{sec.methods}
In Subsection~\ref{sec.building.blocks}, three building blocks for the approaches presented in Subsection~\ref{sec.approaches} are described.

\subsection{Building Blocks}\label{sec.building.blocks}

\textit{Stability Measures:} 
Let $V_1, \ldots, V_m$ denote $m$ sets of selected features, $\labs V_i \rabs$ the cardinality of set $V_i$, and $E\left[\cdot\right]$ the expected value for a random feature selection.
The unadjusted stability measure SMU and the adjusted stability measure SMA (originally called SMA-Count in \cite{bommert2020adjusted}) are defined as
\begin{align*}
&\text{SMU} = \frac{2}{m (m-1)} \sum\limits_{i=1}^{m-1} \sum\limits_{j = i+1}^m \frac{\labs V_i \cap V_j \rabs - E\left[ \labs V_i \cap V_j \rabs\right]}{\sqrt{ \labs V_i \rabs \cdot \labs V_j \rabs} - E\left[ \labs V_i \cap V_j \rabs \right]},\\
&\text{SMA} = \frac{2}{m (m-1)} \sum\limits_{i=1}^{m-1} \sum\limits_{j = i+1}^m \frac{\labs V_i \cap V_j \rabs + \text{Adj}(V_i, V_j) - E\left[ \labs V_i \cap V_j \rabs + \text{Adj}(V_i, V_j)\right]}{\sqrt{ \labs V_i \rabs \cdot \labs V_j \rabs} - E\left[ \labs V_i \cap V_j \rabs + \text{Adj}(V_i, V_j)\right]},\\
&\text{Adj}(V_i, V_j) = \min\{ A(V_i, V_j), A(V_j, V_i)\},\\
&A(V_i, V_j) = \labs \{ x \in (V_i \setminus V_j) : \exists y \in (V_j \setminus V_i) \text{ with similarity}(x, y) \geq \theta\} \rabs.
\end{align*}
The two stability measures show a desirable behavior both on artificial and on real feature sets.
For details on the computation of the stability measures and a thorough comparative study see~\cite{bommert2020adjusted}.

\textit{Stability Selection:}
Stability selection~\cite{meinshausen2010stability,shah2013variable} is a framework for selecting a stable subset of features.
It can be combined with any feature selection algorithm for which the number of features to choose can be set.
It repeatedly applies the feature selection algorithm on subsamples of a given data set and finally selects the features that have been selected for sufficiently many subsamples.

\textit{$\epsilon$-Constraint Selection:}
Having obtained a Pareto front, it often is desirable to choose one point from the front, that provides a good compromise between the objectives, in an automated way.
In the following, we present the algorithm \enquote{$\epsilon$-constraint selection} for choosing such a point for the bi-objective maximization problem of finding a configuration with maximal predictive accuracy and maximal feature selection stability.
It implements an a posteriori $\epsilon$-constraint scalarization method~\cite{miettinen2008introduction}.
(1) Determine maximal accuracy \textit{acc.max} among all configurations.
(2) Remove all configurations with accuracy $< \textit{acc.max} - \textit{acc.const}$.
(3) Among the remaining configurations, determine the maximal stability \textit{stab.max}.
(4) Remove all configurations with stability $< \textit{stab.max} - \textit{stab.const}$.
(5) Among the remaining configurations, determine the maximal accuracy \textit{acc.end}.
(6) Remove all configurations with accuracy $<$ \textit{acc.end}.
(7) If more than one configuration is left, then determine the maximal stability \textit{s.end} among the remaining configurations.
(8) Remove all configurations with stability $<$ \textit{s.end}.
(9) If more than one configuration is left, then randomly choose one of the remaining ones.

\subsection{Proposed Approach and Competing Approaches}\label{sec.approaches}
Our goal is finding models with high predictive accuracy that include all relevant information for target prediction but no irrelevant or redundant features.
For classification data sets with similar features, we propose the approach \enquote{adj}.
We compare it to the three competing approaches \enquote{unadj}, \enquote{acc}, and \enquote{stabs}.
\begin{sloppypar}
\begin{enumerate}[leftmargin = 1.75cm]
\item[\enquote{adj}:] Use $L_0$-regularized logistic regression as predictive method.
$L_0$-regu\-larized logistic regression has one hyperparameter that balances the goodness of the fit and the sparsity of the model.
Tune this hyperparameter with respect to predictive accuracy and to feature selection stability.
This way, a set of Pareto optimal configurations is obtained.
Choose the best configuration with $\epsilon$-constraint selection.
For assessing the stability of the feature selection during hyperparameter tuning, employ an adjusted stability measure.
\item[\enquote{unadj}:] Proceed as in \enquote{adj}, but employ an unadjusted stability measure instead of an adjusted measure.
\item[\enquote{acc}:] $L_0$-regularized logistic regression with hyperparameter tuning only w.r.t.\ predictive accuracy. 
Single-criteria hyperparameter tuning w.r.t.\ predictive accuracy is the standard approach and serves as baseline.
Either a single best configuration or a set of configurations with the same predictive accuracy on the training data is obtained.
In the latter case, one of these configurations is chosen at random.
\item[\enquote{stabs}:] Perform feature selection with stability selection.
Then fit an unregularized logistic regression model including the selected features.
For stability selection, employ $L_0$-regularized logistic regression as feature selection method and tune the hyperparameters of stability selection with respect to predictive accuracy.
With this approach, either a single best configuration or a set of configurations with the same predictive accuracy on the training data is obtained.
In the latter case, one of these configurations is chosen at random.
\end{enumerate}
\end{sloppypar}

\section{Experimental Results on Simulated Data}\label{sec.exp.sd}
First, the approaches are compared on simulated data.
On simulated data, it is known which features have been used for creating the target variable and therefore should be selected and included in a predictive model.

\subsection{Experimental Setup}\label{sec.sd.setup}
\textit{Data Sets:}
For data set creation, a covariance matrix $\Sigma$ is defined and then, the data is drawn from the multivariate normal distribution $\mathcal{N}(0, \Sigma)$.
The covariance matrices considered in this analysis have a block structure.
The features within a block all have Pearson correlation 0.95 to each other and 0.1 to features that are not in this block.
All features have unit variance, making the covariance matrices equal to the respective correlation matrices.
The features within a block are interpreted as similar to each other. 
Given the data, five features $X_1, \dots, X_5$ from different blocks are chosen as explanatory variables.
Then, the class variable $Y_i$ is sampled from a Bernoulli distribution with probability $P(Y_i = 1) = \frac{\exp(\eta_i)}{1 + \exp(\eta_i)}$ with $\eta_i = x_{1,i} + x_{2,i} + x_{3,i} + x_{4,i} + x_{5,i}$, for $i = 1, \dots, n$.
We consider 12 simulation scenarios defined
by all possible combinations of number of observations $n = 100$, number of features $p \in \{ 200, 2\,000, 10\,000 \}$, and block size $\in \{ 1, 5, 15, 25 \}$.

\textit{Setup for Hyperparameter Tuning:}
For $L_0$-regularized logistic regression the hyperparameter indicating the (maximum) number of features to be included in the model needs to be tuned.
Because in the implementation in the R~package \textit{L0Learn} this hyperparameter is of type integer, a grid search for the best value is performed.
To evaluate the performance of each hyperparameter value, 10-fold cross-validation is conducted.
For \enquote{acc}, the mean classification accuracy of the 10~models on the respective left-out observations for the 10~cross-validation iterations is assessed.
For \enquote{adj} and \enquote{unadj}, the mean classification accuracy and the feature selection stability are evaluated.
The feature selection stability is quantified based on the 10~sets of features that are included in the 10~models.
In the \enquote{adj} approach, the adjusted stability measure SMA is employed for stability assessment.
SMA interprets features from the same block as exchangeable.
In the \enquote{unadj} approach, the unadjusted stability measure SMU is used.
The values of the stability measures are calculated with the R~package \textit{stabm}.
Based on the performance values, the best configuration is selected.
For $\epsilon$-constraint selection, the cutoff values, $\textit{acc.const} = 0.025$ and $\textit{stab.const} = 0.1$ are used.
These values have been determined in preliminary studies.
For \enquote{stabs}, the implementation of stability selection from the R~package \textit{stabs} is used with 50 complementary subsamples.
Two real-valued hyperparameters are tuned with a random search.
To determine the quality of hyperparameter values, the classification accuracy of an unregularized logistic regression model with the selected features is evaluated using 10-fold cross-validation.

\textit{Evaluation:}
For each approach, a final model is built based on the entire data set.
For \enquote{adj}, \enquote{unadj}, and \enquote{acc}, a $L_0$-regularized logistic regression model with the best hyperparameter value is fitted.
For \enquote{stabs}, stability selection is conducted with the best hyperparameter values.
Then, an unregularized logistic regression model is built with the selected features.
Additionally, an unregularized logistic regression model with the five features that were used for generating the target variable is fitted.
This provides an upper bound for the predictive accuracy that can be achieved and will be denoted by \enquote{truth}.
Based on the final models, three performance measures are calculated: the classification accuracy on new test data, the number of false positive features, and the number of false negative features.
The number of false positive features is the number of irrelevant or redundant features that have been selected for the final model.
The number of false negative features is the number of relevant and not redundant features that have not been selected for the final model.
For the assessment of the number of false positive and false negative features, features from the same block are interpreted as exchangeable.
So, if instead of a feature that was used for generating the target variable, an other feature from the same block is selected, this other feature is accepted as well.
For evaluating the test accuracy, a new test data set of the same size is created in the same way as the training data set.
Then, the classification accuracy of the final models is assessed.
To ensure a fair comparison, all approaches use the same training and test data sets as well as the same cross-validation splits.
For each simulation scenario, 50~training and test data sets are created.

\subsection{Results}

\begin{figure}[p]
\centerline{\includegraphics[width = \textwidth, page = 1]{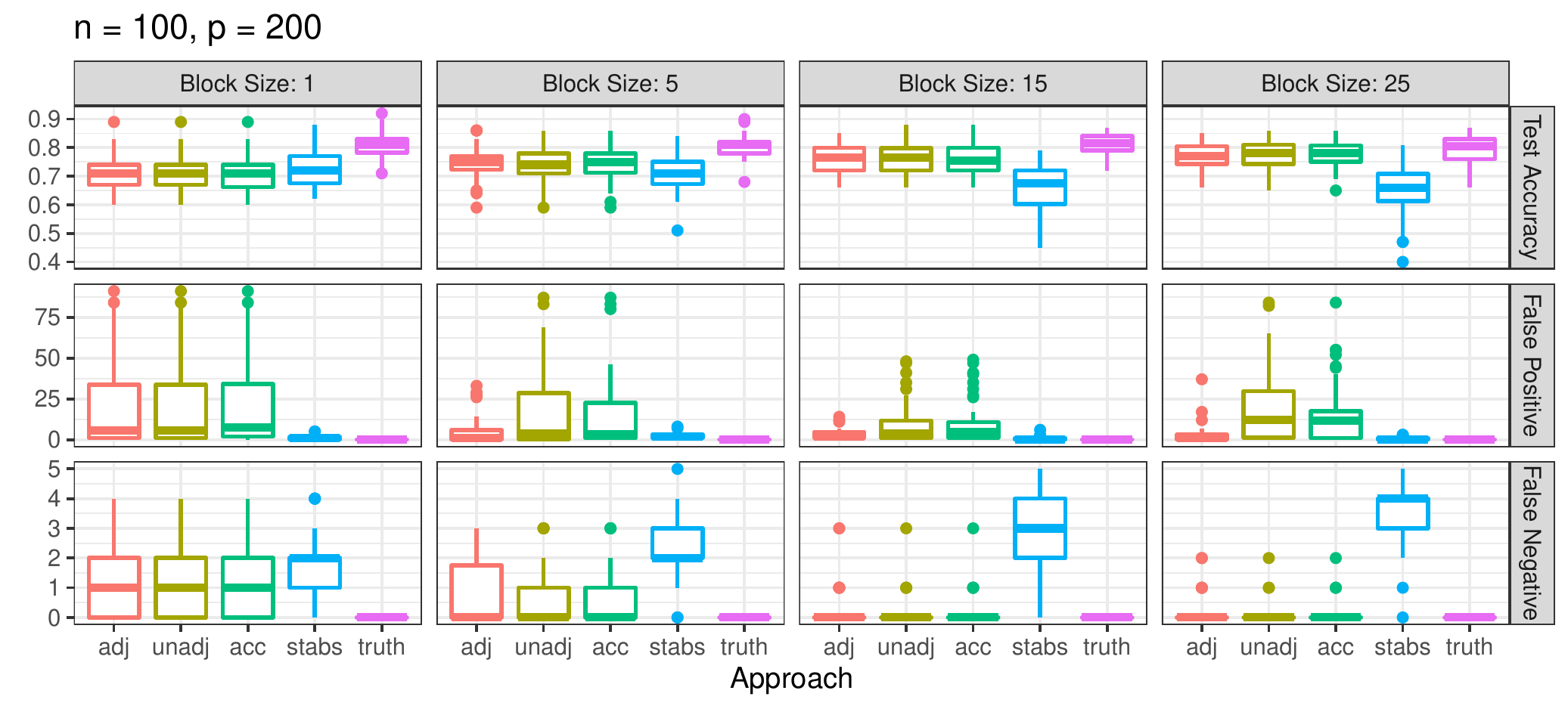}}
\centerline{\includegraphics[width = \textwidth, page = 2]{sd_results.pdf}}
\centerline{\includegraphics[width = \textwidth, page = 3]{sd_results.pdf}}
\caption{\enquote{Test Accuracy}: classification accuracy on independent test data.
\enquote{False Positive}: number of irrelevant or redundant features that have been selected.
\enquote{False Negative}: number of relevant and not redundant features that have not been selected.}
\label{fig.sd}
\end{figure}

The results of the simulation study are shown in Fig.~\ref{fig.sd}.
First, the scenarios with similar features are analyzed.
In these scenarios, the predictive performances of the models obtained with \enquote{adj}, \enquote{unadj}, and \enquote{acc} are very similar.
In the scenarios with $p=200$, the classification accuracies are quite close to the upper bound: the classification accuracy of a model that employs exactly the five features used for target generation.
In the scenarios with $p=2\,000$ and $p=10\,000$, the predictive performances of the models resulting from the three approaches are noticeably lower than the upper bound.
But, the more similar features there are, the closer are the predictive performance values to the upper bound.

Using the adjusted stability measure during tuning leads to much fewer false positive features compared to single-criteria tuning and to tuning using the unadjusted stability measure.
So, the models obtained with \enquote{adj} contain fewer irrelevant or redundant features than the models resulting from \enquote{unadj} and \enquote{acc}.
This advantage comes at the small drawback of a slightly increased number of false negative features which, however, does not result in a decreased predictive performance.
In the situations with $p=200$, these observations can be made for all simulation scenarios with similar features.
In the settings with $p=2\,000$ or $p = 10\,000$, blocks of size 15 or 25, respectively, are necessary.
It should be noted that in the high-dimensional simulation scenarios, the relative number of similar features is much lower than in the low-dimensional settings.

In the scenarios with similar features, \enquote{stabs} performs worse than the other approaches in terms of predictive accuracy.
The low classification accuracy is due to too few relevant features being selected: the number of false negative features is high.
The models obtained with \enquote{stabs} are very sparse.
They contain almost no irrelevant or redundant features, but also not many relevant features.

When there are no similar features, all approaches lead to models with similar classification accuracy.
With the stability selection approach, the resulting models contain the fewest irrelevant or redundant features among the compared approaches.
In the scenarios without similar features, the proposed approach does not perform worse than the standard approach \enquote{acc}, even though the proposed approach was specifically designed for situations with similar features.
In the situation with no similar features, the approaches \enquote{adj} and \enquote{unadj} are identical because in this situation, the two stability measures are identical.

Now, we consider the influence of the simulation scenarios on the performance of each approach.
For the approaches \enquote{adj}, \enquote{unadj}, and \enquote{acc}, the predictive accuracy increases with increasing block size.
The larger the blocks of similar features, the more features are similar to the 5~features used for target generation.
So, it becomes easier to select features with information for target prediction.
In all scenarios, for \enquote{adj}, the number of false positive features decreases with increasing block size.
The larger the blocks of similar features, the fewer irrelevant and the more redundant features there are in the data set.
So, it can be concluded that with \enquote{adj}, especially the selection of redundant features is prevented.
For the other approaches, an analogous decrease in the number of false positive features cannot be observed.
For \enquote{adj}, \enquote{unadj}, and \enquote{acc}, in almost all scenarios, the number of false negative features decreases with increasing block size.
The larger the blocks of similar features, the more relevant features there are in the data set, making it easier for the methods to select relevant features.

The prediction accuracy of the \enquote{stabs} approach decreases with an increasing number of false negative features.
The number of false negative features increases with increasing block size in the scenarios with $p=200$.
Recall that $L_0$-regularized logistic regression usually selects only one feature out of a group of similar features.
When repeatedly performing feature selection on the subsamples, it is likely that each time, only one feature out of each group of similar and relevant features is selected and that the selection frequencies within a group are fairly equal. 
So, if the blocks are large, the selection frequencies become very small.
If the highest of the selection frequencies is below any reasonable value of the inclusion threshold, none of the features is included in the final model.
In the scenarios with $p=2\,000$ and $p=10\,000$, the number of false negative features is almost constant with respect to the number of similar features.
Comparably many relevant and not redundant features are not included in the final models.

\section{Experimental Results on Real Data}\label{sec.exp.trd}
Now, the approaches are compared based on 12~real data sets from the platform \textit{OpenML}~\cite{vanschoren2013openml} and the R~package \textit{datamicroarray}.
The data sets come from various domains and differ in dimensions and in feature similarity structures.

\subsection{Experimental Setup}
In this study, whenever possible, tuning and evaluation are performed in the same way as in Section~\ref{sec.exp.sd}.
Because it is not possible here to generate independent test data sets, nested cross-validation with 10~inner and 10~outer iterations is used.
The inner iterations are used to determine the best configurations and the left-out observations of the outer iterations are used to evaluate the predictive accuracy.
So, for each approach and each data set, 10~predictive performance values are obtained.
The number of false positive and false negative features cannot be assessed.
Instead, the number of features included in the 10~models, whose predictive accuracy is evaluated on the left-out test data, is recorded.

\subsection{Results}
\begin{figure}[p]
\centerline{\includegraphics[width = \textwidth, page = 1]{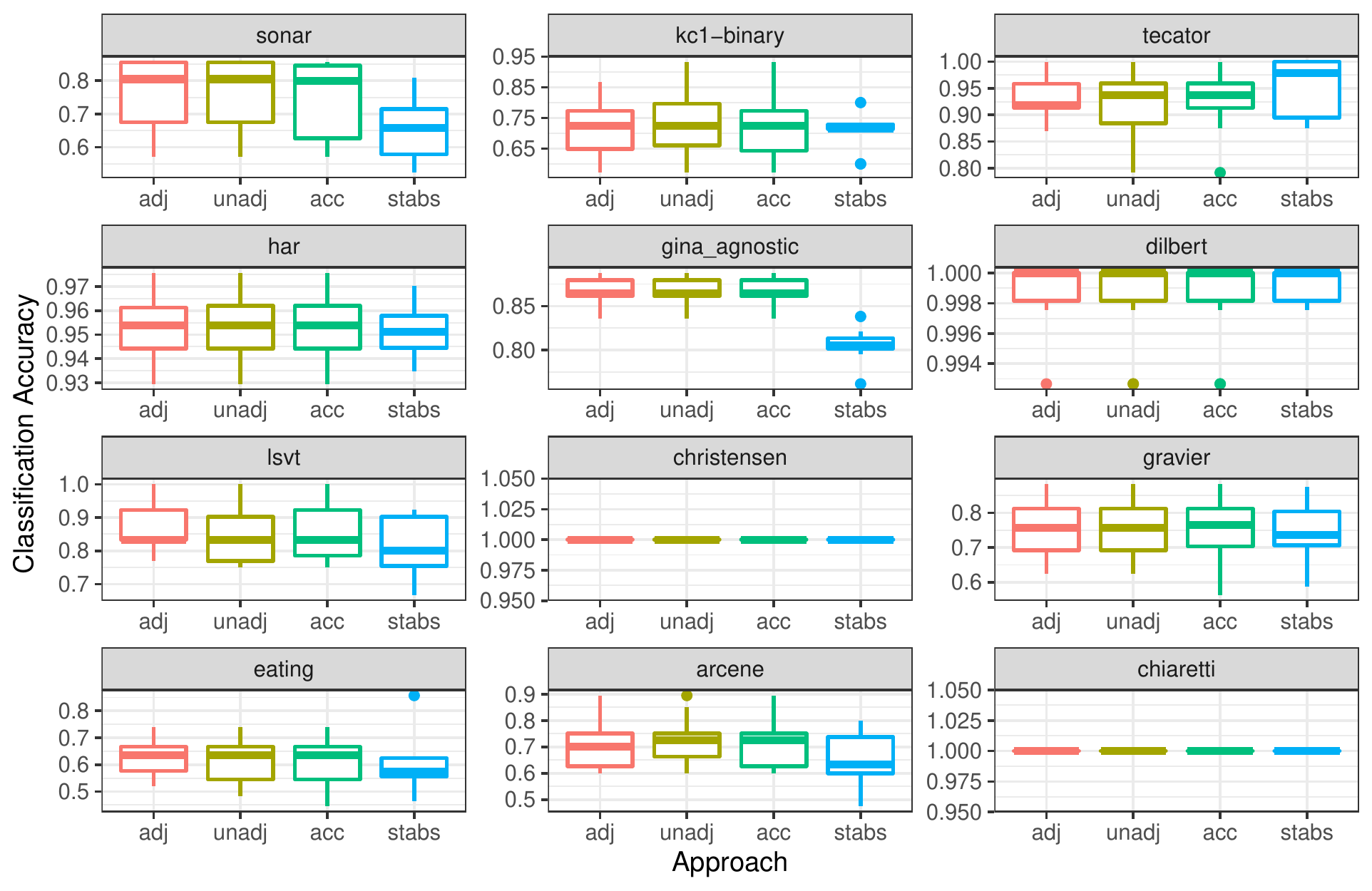}}
\vspace*{0.25cm}
\centerline{\includegraphics[width = \textwidth, page = 2]{trd.pdf}}
\caption{Top: Classification accuracy of the best configurations on the left-out test data of the outer cross-validation iterations.
Bottom: Number of selected features of the best configurations per outer cross-validation iteration.}
\label{fig.trd}
\end{figure}

The top plot in Fig.~\ref{fig.trd} shows the classification accuracy of the best configurations.
The bottom plot displays the number of features that are selected for these configurations.
For data sets \textit{sonar}, \textit{tecator}, \textit{har}, \textit{dilbert}, \textit{lsvt}, \textit{christensen}, \textit{arcene}, and \textit{chiaretti}, it is beneficial to perform multi-criteria tuning with respect to both classification accuracy and stability and choosing the best configuration based on $\epsilon$-constraint selection, compared to single-criteria tuning only with respect to classification accuracy.
A comparable or even better predictive performance is achieved with multi-criteria tuning and the fitted models include fewer features.
Among these data sets, for \textit{tecator}, \textit{har}, \textit{dilbert}, \textit{lsvt}, and \textit{arcene}, it is necessary to use the adjusted stability measure to achieve this benefit.
These data sets contain many similar features.
For data sets \textit{sonar}, \textit{christensen}, and \textit{chiaretti}, which contain only few similar features, the unadjusted stability measure is sufficient.
For data sets \textit{kc1-binary}, \textit{gina\_agnostic}, \textit{gravier}, and \textit{eating}, multi-criteria tuning does not provide a benefit over single-criteria tuning.
Still, for these data sets considering the feature selection stability during tuning does not decrease the predictive performance or increase the number of selected features of the resulting models in comparison to single-criteria tuning.

Comparing the results of the proposed approach and \enquote{stabs}, the proposed approach performs better on the majority of data sets.
For data sets \textit{sonar}, \textit{har}, \textit{gina\_agnostic}, \textit{lsvt}, \textit{gravier}, \textit{eating}, and \textit{arcene}, stability selection leads to a worse predictive accuracy than the other approaches.
For data sets \textit{christensen} and \textit{chiaretti}, it fails at excluding irrelevant or redundant features.
Only for data set \textit{tecator}, models with higher predictive accuracy are obtained with stability selection compared to the other approaches.
For \textit{kc1-binary} and \textit{dilbert}, more sparse models with the same predictive quality can be fitted with the stability selection approach.
On most data sets, the stability selection approach leads to comparably sparse models, often at the expense of a comparably low predictive accuracy.
This has been observed on simulated data as well.

\section{Conclusions}\label{sec.conclusion}
Fitting models with high predictive accuracy that include all relevant but no irrelevant or redundant features is particularly challenging for data sets with similar features.
We have proposed the approach of tuning the hyperparameters of a predictive model in a multi-criteria fashion with respect to predictive accuracy and feature selection stability.
We have used $L_0$-regularized logistic regression as classification method in our analysis because it performs embedded feature selection and -- in contrast to many state-of-the-art feature selection methods -- is able to select only one feature out of a group of similar features in a data set.
We have evaluated the approach of multi-criteria-tuning its hyperparameter based on simulated and on real data.
We have compared it to the standard approach of single-criteria tuning of the hyperparameter as well as to the state-of-the-art technique stability selection.

On simulated data, especially in the scenarios with many similar features, tuning the hyperparameter of $L_0$-regularized logistic regression with respect to both predictive accuracy and stability is beneficial for avoiding the selection of irrelevant or redundant features compared to single-criteria tuning. 
To obtain this benefit, the feature selection stability must be assessed with an adjusted stability measure, that is, a stability measure that considers similarities between features.
Measuring the stability with an unadjusted measure does not outperform single-criteria tuning in most scenarios.
Also, considering the stability during tuning does not decrease the predictive accuracy of the resulting models.

On real data, performing hyperparameter tuning with respect to both predictive accuracy and feature selection stability can be beneficial for fitting models with fewer features without losing predictive accuracy.
For data sets with many similar features, an adjusted measure must be used while for data sets with only few similar features, an unadjusted measure is sufficient.
For all data sets, almost no predictive accuracy is lost by additionally considering the stability.

Compared to the stability selection approach, models with higher predictive accuracy are fitted with the proposed approach, especially in simulation scenarios with many similar features.
On real data sets, the proposed approach outperforms stability selection on many of the data sets.
Both on simulated data and on real data, with the stability selection approach, comparably sparse models are fitted.
These models, however, often do not include enough relevant features and therefore obtain a comparably low predictive accuracy.
Also, the stability selection approach takes much more time for computing.
For larger data sets, it is infeasible without a high performance compute cluster.

The proposed approach is not only applicable to classification data but also to regression or survival data by using the respective $L_0$-regularized methods.
Also, other feature selection methods that are able to select only one feature out of a group of similar features can be used instead of $L_0$-regularized methods.

\section*{Acknowledgments}
This work was supported by German Research Foundation (DFG), Project RA\,870/7-1 and Collaborative Research Center SFB~876, A3. We acknowledge the computing time provided on the Linux HPC cluster at TU Dortmund University (LiDO3), partially funded in the course of the Large-Scale Equipment Initiative by the German Research Foundation (DFG) as Project 271512359.

\bibliographystyle{splncs04}
\bibliography{References}
\end{document}